# Performance of Transfer Learning Model vs. Traditional Neural Network in Low System Resource Environment

William Hui, Active Intelligence Holdings Limited

## Abstract

Recently, the use of pre-trained model to build neural network based on transfer learning methodology is increasingly popular. These pre-trained models present the benefit of using less computing resources to train model with smaller amount of training data. The rise of state-of-the-art models such as BERT, XLNet and GPT boost accuracy and benefit as a base model for transfer leanring. However, these models are still too complex and consume many computing resource to train for transfer learning with low GPU memory. We will compare the performance and cost between lighter transfer learning model and purposely built neural network for NLP application of text classification and NER model.

## 1. Introduction

Transfer learning model is based on a layer of pre-trained deep learning multi-layer model ("Base Model") connected to a simple customed built layer ("Fine Tuning Layer"). The Base Model is a general purpose pre-trained model that can be used for any application. For the scope of NLP, the state-of-art models include BERT, XLNet and GPT. Some of these models can also be used for image recognition purpose. The Base Model is pre-trained for a task within the same domain with large amount of data. In the context of NLP, the Base Model is trained for one task such as predicting the next word of sentence to generate converged word embedding. The Base Model can be fine tuned to perform another task in the same domain such as text classificaiton. Usually the fine tuning layer is much simpler and require less data to train.

The advantage of transfer learning is that these pre-trained models are available and ready to use. The simplicity and less data requirement of fine tuning layer save much computing power for training and effort to label training data.

According to *A survey on Transfer Leanring* (Pang and Yang, 2019), transfer learning may not yield better or same accurcy as standalone machine learning model.

Despite the saving in computing resources, some models such as BERT which the BERT large uncase model from Tensorflow Hub consists of 24-layer, 1024 hidden node, 16 head and 340 million parameters, making it difficult to train with tiny GPU even for transfer learning.

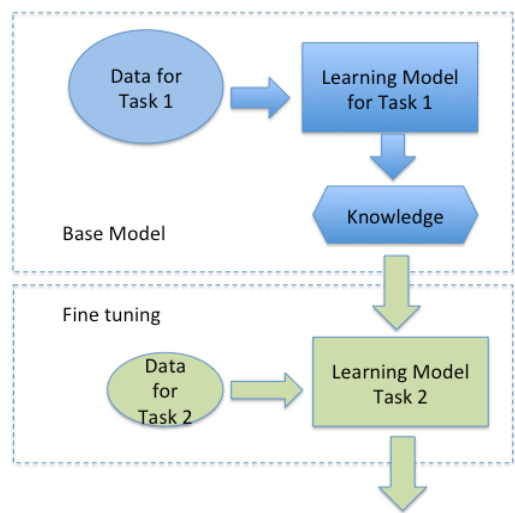

Fig 1: Transfer Learning Architecture

In this paper we will study the transfer learning model using lighter Base Models and compare their performance with the conventional neural network with more customized features.

We will use the model to classify short paragraphs of financial news to identify whether the article is about merger and acquisition, M&A, fund raising or just general news. The article may contain company names that cannot be identified by pre-trained word embedding, financial figures in various format and terms used in investment banking such as "IPO", "Series A"

## 2. Related Work

**Word Embedding:** Sentence is divided into a series of words, each word is encoded to numerical vector. Earlier word embedding methodology such as word2vec (Goldberg, Levy, 2014) and GloVe (Pennington, Socher, Manning, 2014) provide one-to-one vocabulary-to-vector relationship which does not account for more than one meaning of a single word in different sentence context. In addition, word embedding model recognizes strange word such as names and symbol as unkown vector which will lose the meaning in the sentence. ELMO (Peters, Neumann, Iyyer Gardner, Clark, Lee, Zettlemoyer , 2018) considers the meaning of the word in the sentence and produce different vectors for the same word in different sentence context. However, ELMO model must work in a fixed length structure, which affects the efficiency and loses some of context for long sentence. Word embedding are usually inputted to an RNN to produce a sentence context. The RNN also suffers from vanishing gradient for long sentences.

**Sentence Embedding:** Sentence embedding model encodes the sentence or paragraph and produces a sentence embeding vector. Bi-directional Encoder Representations from Transformer, or BERT (Devlin, Chang, Lee, Toutanova, 2018) is one of the most robust model as a pre-trained base model. However, its multi-layers and high number parameters architecture makes it hard to run even as a pre-trained model with moderate GPU. Universal Sentence Encoder (Cer, Yang, Kong, Hua, Limtiaco, St. John, Constant, Guajardo-Cespedes, Yuan, Tar, Sung, Strope, Kurzweil, 2018) is a lighter alternative that can handle simple task equally well in a small hardware.

**Attention based model:** Attention model addresses the issue of vanishing gradient by considering different weighting from each word to the sentence context. Similar to word embedding, if the upstream word embedding is not recognizing rare and unconventional words such as company name, it will limit the performance of the attention model.

## 3. Approach

To evaluate the performance of transfer learning model against the conventional neural network in a constrained resource environment, few models are built to classify financial news with headline and a synopsis of the news up to 3 sentences and classify whether the news is related to M&A, equity raising or just a general news.

```
<headline>Cellular  Dynamics
Raises $40.6 million</headline>
```

```
<Synopsis> the  world's highest
volume  manufacturer  of  human
```

heart cells, has closed on a $40.6 million Series B private equity round. This financing enables the company to, human heart cells derived from induced pluripotent stem cells (iPSCs), and to launch additional human tissue cell products for biomedical and pharmaceutical drug development and safety research. CDI also plans to use the proceeds to rapidly expand its commercial organization to meet the growing demand for these iPSC-based products. CDI has raised a total of $70 million since 2004. </Synopsis>

Exhibit #2: Sample News Data

## 4. Classification Models

Few models are evaluated for text classification:

### 4.1 Transfer learning model with ELMO

Word embedding based transfer learning model with ELMO and 1 dense layer as the fine tuning layer. The entire news articles are trimmed or padded to 50 words and then passed to ELMO to generate N x 60 x 1024 vectors. The vectors are then passed to a dense layer and a softmax layer to classify the news type. ELMO produces high level of accuracy for articles for strange words such as non-English company name. However, the accuracy is lower for company names which are also legitimate English words.

### 4.2 Universal Sentence Encoder based Model

Sentence embedding based transfer learning model is built using Universal Sentence Encoder from Tensorflow Hub as the base model with a dense layer and a softmax layer to classify news type. The Universal Sentence Encoder and reads the entire sentence and paragraph as a string and produces a N x 512 sentence vectors. The fine tuning layer consisting of dense layer and softmax layer are then applied to the vectors to classify the news type. The simplicity of the model allows fast training.

### 4.3 Sentence embedding with attention layer

The synopsis and the title are tokenized into three sentences. Each sentence is embedded by the Universal Sentence Encoder model separately and produces a N x 3 x 512 vectors. The vectors are then passed to a self-attention model to calculate the weight of each sentence to the end classification result.

### 4.4 Conventional Recurrent Neural Network with customized features

In contrast to the transfer learning model, a conventional RNN is built with additional features that captures other attribute of the word in addition to the tokenization of the word itself. The model is based on similar model that extracts key information from the news including names of parties involved, transaction amount and investment stage.

During the preprocessing, additional work is performed to set features:

- The words are tokenized to seprate the prefix or suffix of the word. The word "re-construct" is divided into two words: "re" and "construct".
- For unknown words, if the word is not an legitimate english word, the word is tagged as 'ILGTE' versus 'UNK' for legitimate word.
- Numbers with dollar sign are tagged differently than other numbers such as date and quantifier.

The conventional word embedding + GLOVE + RNN model does not address a unique characteristics of the news article that many strange vocabularies such as company name and amount appear very rarely. These vocabularies have no GloVe embedding vector representation. It is hard for word embedding to converge if these vocabularies only appear once or twice in large number of training data.

To compensate the deficiency of non-GloVe word, we add a few word context to the model that are lost in the word embedding. The following word context features are added to the model:

- Does it start with a capital letter?
- Is the word a legitimate english word?

- The Part of Speech Tag, POS (Bird, Klein and Loper, 2009)
- Does the word contain ".com" or ".ai"

The word context has extra relationship between the word in relation to its neighbouring words.

The POS tag is also embedded to a vector to correlate the relationship between tags. The model reads input word tokens of the article and the respective word context feature. The word tokens are embedded and then passed to a bi-directional GRU. On the other hand, the POS is embeded to produce embedding vector. The vector is merged with other word context vector and then passed to another bi-directional GRU.

The final hidden states of the two bi-directional GRU is merged and passed to a fully connected layer followed by a softmax layer to predict the probability of each transaction type.

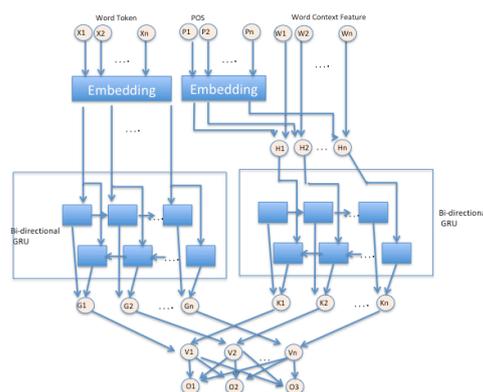

Exhibit 3: Conventional Recurrent Neural Network with purpose built features

GRU is preferred over LSTM as the test reported in the paper (Gruber, Jockisch 2020) shows GRU cells tend to learn content that is rarely found in the data (low prevalence) better than LSTM cells, which fits our constraint of limited and scattered training data content.

## 5. Experiment Condition

The corpus contains 14,400 articles. Table 4 outlines the detail of the corpus

|  | Training | Validation | Test |
|---|---|---|---|
| Articles | 6,332 | 1,809 | 905 |
| Sentences | 18,070 | 5,894 | 2,931 |
| Words | 440,445 | 144,624 | 72,311 |
| Illegitimate words | 171,820 | 56,418 | 28,210 |
| No of Fund raising News (label) | 2885 | 824 | 412 |
| No of M&A News (label) | 522 | 149 | 75 |
| No of General News News (label) | 2925 | 836 | 418 |

Table 4: Statistics of the corpus

The hardware used to train the model is Intel I3 CPU with 8GB RAM and NVIDIA GeForce 1050 GPU with 3GB memory.

Tensorflow's Keras' default word embedding library is used. The pre-trained model of ELMO verion 2 and Universal Sentence Encoder version 2 from Tensorflow Hub are used. For fine tuning data, 3 years (2018 – 2020) of capital market news articles in health care, fintech industries are used for pre-training. The fine tuning layer consists of 32 units of fully connected layer followed by 3 units of softmax layers.

For the Conventional Recurrent Neural Network with purpose built features ("RNN-Plus"), GloVe is used as default embedding, 50 time steps are used for the Bi-RNN. 30 units of fully connected layer with Relu activation is used for both word token layer and word context layer after their respective Bi-RNN. A softmax layer is then applied after the results of the fully connected layer are merged.

The output of the model will show the classification result whether it is M&A news, fund raising news or general news.

## 6. Results

The result of the four models are evaluated based on accuracy and F1 score.

The purposed built RNN-GloVe plus shows higher accuracy than transfer learning models.

| Model | Accuracy |
|---|---|
| ELMO+Dense | 81.46% |
| USE+Dense | 84.60% |
| USE+Attention | 83.28% |
| RNN Plus | **91.77%** |

Table 5: Accuracy

We also measure the precision, recall and the F1 score.

|  | General News | Fund Raising News | M&A News |
|---|---|---|---|
| Precision | **0.9053** | **0.9363** | **0.8730** |
| Recall | **0.9233** | 0.9233 | **0.8462** |
| F1 | **0.9142** | **0.9297** | **0.8594** |

Table 6: RNN Plus

|  | General News | Fund Raising News | M&A News |
|---|---|---|---|
| Precision | 0.7726 | 0.8420 | 0.7947 |
| Recall | 0.8492 | 0.7939 | 0.6240 |
| F1 | 0.8091 | 0.8172 | 0.6991 |

Table 7: ELMO+Dense

|  | General News | Fund Raising News | M&A News |
|---|---|---|---|
| Precision | 0.7846 | 0.875 | 0.6363 |
| Recall | 0.8226 | 0.875 | 0.4118 |
| F1 | 0.8031 | 0.875 | 0.5 |

Table 8: USE+Dense

|  | General News | Fund Raising News | M&A News |
|---|---|---|---|
| Precision | 0.5556 | 0.5336 | 0.625 |
| Recall | 0.0229 | **0.9838** | 0.6522 |
| F1 | 0.0441 | 0.6919 | 0.6383 |

Table 9: USE+Attention

As the training data has less M&A labelled data than the other two labels, the accuracy and F1 score for M&A news is lower than the other two across all models. Overall, the purpose built RNN-Plus model outperforms the transfer learning model when the volume of training data is more limited.

Among the transfer learning model, USE+Dense outperforms the other two transfer learning models when there is more training data.

To further evaluate the effectiveness of the word context, we also build a character embedding for the non-English word to create a word embedding rather than using the [ILEGTE] or [UNK]. The improvement is not sigificant.

## 7. Conclusion

Models built from transfer learning demonstrates decent accuracy. However, under the condition of small dataset, the purpose built model outperforms transfer learning.

As an alternative of BERT under constrained GPU resources, between sentence embedding based USE and word embedding based ELMO which factors sentence context, USE is slightly better than ELMO. But ELMO has better accuracy when dataset is small.

The main advantage of the RNN Plus over the transfer learning is that RNN Plus reads the feature of POS and other word context for unknown words such as company name and product name. The model combines the results from word context pattern and word embedding sequence versus transfer learning which only learns from the word or sentence embedding. The result shows having additional context features when some critical keywords are unknown helps increase accuracy of the model and enhance the recall number.

With limited GPU resources, transfer learning shall equip with sufficient dataset. Although the performance is slightly worse than customed built neural network, the model can be quickly constructed to use while we study the engineering of customed features for more accurate customed built model.